\documentclass[letterpaper, 10 pt, conference]{ieeeconf}

\IEEEoverridecommandlockouts                              

\usepackage{graphics} 
\usepackage{epsfig} 
\usepackage{times} 

\usepackage{gensymb}

\usepackage{cite}
\usepackage{textcomp}
\usepackage{amsmath,amssymb,amsfonts}
\usepackage{multicol}
\usepackage{float} 
\usepackage{tablefootnote}
\usepackage{url}
\usepackage{hyperref}
\usepackage{subfigure}
\usepackage{multirow}
\usepackage{array}
\usepackage{algorithm}
\usepackage{algpseudocode}
\usepackage{xcolor}
\usepackage{booktabs}
\usepackage{comment}
\newcolumntype{P}[1]{>{\centering\arraybackslash}p{#1}}

\usepackage{scalerel}
\usepackage{tikz}
\usetikzlibrary{svg.path}

\definecolor{orcidlogocol}{HTML}{A6CE39}
\tikzset{
  orcidlogo/.pic={
    \fill[orcidlogocol] svg{M256,128c0,70.7-57.3,128-128,128C57.3,256,0,198.7,0,128C0,57.3,57.3,0,128,0C198.7,0,256,57.3,256,128z};
    \fill[white] svg{M86.3,186.2H70.9V79.1h15.4v48.4V186.2z}
                 svg{M108.9,79.1h41.6c39.6,0,57,28.3,57,53.6c0,27.5-21.5,53.6-56.8,53.6h-41.8V79.1z M124.3,172.4h24.5c34.9,0,42.9-26.5,42.9-39.7c0-21.5-13.7-39.7-43.7-39.7h-23.7V172.4z}
                 svg{M88.7,56.8c0,5.5-4.5,10.1-10.1,10.1c-5.6,0-10.1-4.6-10.1-10.1c0-5.6,4.5-10.1,10.1-10.1C84.2,46.7,88.7,51.3,88.7,56.8z};
  }
}

\newcommand\orcidicon[1]{\href{https://orcid.org/#1}{\mbox{\scalerel*{
\begin{tikzpicture}[yscale=-1,transform shape]
\pic{orcidlogo};
\end{tikzpicture}
}{|}}}}

\usepackage{graphicx} 

\normalmarginpar 
\newcommand{\IEEEpreprintnotice}{
    \rotatebox{90}{\footnotesize This work has been submitted to the IEEE for possible publication. 
    Copyright may be transferred without notice, after which this version may no longer be accessible.}
}

\title{\LARGE \bf \vspace{-10pt} 
CARIL: Confidence-Aware Regression in Imitation Learning for Autonomous Driving
}

\author{Elahe Delavari$^{1}$\orcidicon{0009-0001-8768-7903}, Aws Khalil$^{2}$\orcidicon{0000-0001-9139-3900}, and Jaerock Kwon$^{3}$\orcidicon{0000-0002-5687-6998}
\thanks{*This work was supported in part by the National Science Foundation (NSF) under Grant MRI 2214830. $^{1}$Elahe Delavari, $^{2}$Aws Khalil, and $^{3}$Jaerock Kwon are with the Department of Electrical and Computer Engineering, University of Michigan-Dearborn, Dearborn, MI, USA.
{\tt\small \{elahed, awskh, jrkwon\}@umich.edu}}
}

\begin{document}
    \maketitle
    \marginpar{\IEEEpreprintnotice} 

    \begin{abstract}

End-to-end vision-based imitation learning has demonstrated promising results in autonomous driving by learning control commands directly from expert demonstrations. However, traditional approaches rely on either regression-based models, which provide precise control but lack confidence estimation, or classification-based models, which offer confidence scores but suffer from reduced precision due to discretization. 
This limitation makes it challenging to quantify the reliability of predicted actions and apply corrections when necessary.
In this work, we introduce a dual-head neural network architecture that integrates both regression and classification heads to improve decision reliability in imitation learning. The regression head predicts continuous driving actions, while the classification head estimates confidence, enabling a correction mechanism that adjusts actions in low-confidence scenarios, enhancing driving stability.
We evaluate our approach in a closed-loop setting within the CARLA simulator, demonstrating its ability to detect uncertain actions, estimate confidence, and apply real-time corrections. Experimental results show that our method reduces lane deviation and improves trajectory accuracy by up to $50\%$, outperforming conventional regression-only models. These findings highlight the potential of classification-guided confidence estimation in enhancing the robustness of vision-based imitation learning for autonomous driving.
The source code is available at \url{https://github.com/ElaheDlv/Confidence_Aware_IL}.

\end{abstract}

    \section{Introduction}
\label{sec:introduction}

End-to-end imitation learning has emerged as a powerful paradigm for autonomous driving, allowing vehicles to learn driving policies directly from expert demonstrations \cite{le2022survey}. In these systems, deep neural networks predict control actions, such as steering and acceleration, from raw sensory input. 
However, existing approaches typically rely on either regression-only or classification-only models, each with inherent limitations. 
Regression-based models optimized with Mean Squared Error (MSE) provide precise control predictions but lack a mechanism to estimate confidence in their decisions, making them prone to errors in uncertain or challenging scenarios \cite{cui2019uncertainty,nozarian2020uncertainty}. 
In contrast, classification-based models discretize the action space, offering probabilistic outputs that can represent confidence levels. However, this comes at the cost of reduced precision, ambiguous decision boundaries, and increased computational complexity \cite{stewart2023regression,kishky2024end}. 
These limitations hinder reliable decision-making in dynamic environments.

\begin{figure}[!t]
    \centering
    \includegraphics[width=1\columnwidth, clip, trim=0 14 0 0]{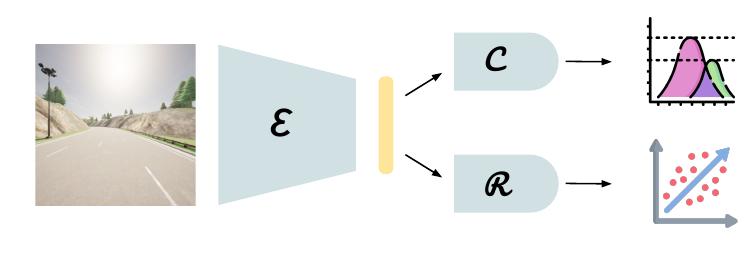}
    \caption{
            Overview of the proposed dual-head architecture for vision-based imitation learning. The model takes a front-facing camera image as input, processed by an encoder ($\varepsilon$) to extract latent features. Two separate heads generate outputs: the classification head ($C$) estimates confidence levels by predicting probability distributions over actions, while the regression head ($R$) predicts continuous control values such as steering and acceleration. The classification output enables real-time confidence assessment, allowing corrective actions when necessary, improving the reliability of imitation learning-based autonomous driving. (Icons are from Flaticon.com)
            }
    \label{fig:method-intro}
\vspace*{-10pt}
\end{figure}

To address these challenges, we propose a novel dual-head neural network architecture that integrates both regression and classification heads to enhance decision reliability in imitation learning-based autonomous driving. Unlike prior multi-task learning approaches that integrate classification for auxiliary tasks \cite{xu_end--end_2017, couto2021generative, kishky2024end}, our model directly leverages classification to quantify action confidence. This enables real-time uncertainty detection and correction, improving overall driving stability. As illustrated in Fig.~\ref{fig:method-intro}, the regression head predicts continuous control commands, while the classification head estimates the confidence of these predictions. This additional confidence estimation enables real-time detection of uncertain predictions, allowing corrective measures to be applied and improving overall driving stability.
This work focuses on two key contributions: (i) We introduce a classification-guided confidence estimation mechanism for imitation learning, enabling real-time detection of uncertain predictions. (ii) We develop an action correction module that adjusts action commands based on the classification head’s confidence output.   

    \section{Related Work}
\label{sec:related-work}

To address the limitations of regression-based imitation learning, researchers have explored uncertainty-aware continuous control models, discretized action spaces, and classification-based multi-task learning frameworks. This section reviews these approaches and their contributions to imitation learning for autonomous driving.

\subsection{Continuous Action space (Regression-Based Learning)}
Regression-based approaches remain the dominant choice in end-to-end imitation learning, where models predict continuous control values directly from sensory inputs \cite{le2022survey}. From early frameworks like ALVINN, Dave, and Dave-2 \cite{pomerleau_alvinn_1989, net-scale_autonomous_2004, chen_end--end_2017} to modern model-based imitation learning \cite{hu2022model, popov2024mitigating}, most methods rely on regression models \cite{bojarski2016end, bojarski2017explaining, codevilla2018ConditionalImitation, wang_end--end_2019-1, wu_end--end_2019, codevilla2019exploring, prakash2021multi, kwon2022incremental, khalil2023anec, kim2024ndst, khalil2024plm}. These models, typically optimized with Mean Squared Error (MSE), enable smooth and precise control but lack confidence estimation, making them prone to errors in occlusions or unseen environments. Moreover, small inaccuracies accumulate over time, leading to Covariate Shift, a well-known issue in imitation learning \cite{ross2011reduction}.

\subsection{Discrete Action Space (Classification-Based Learning)}

While continuous control is often preferred in autonomous driving for its smooth and precise vehicle maneuvers, discrete action spaces have been explored within reinforcement learning (RL) frameworks \cite{kiran2021deep}. Discretizing the action space simplifies decision-making and enables the use of algorithms like Q-learning. For instance, Sallab et al. \cite{sallab2016end} investigate both discrete and continuous action spaces for RL-based autonomous driving strategies. However, this approach can lead to reduced control granularity and potential oscillations in vehicle behavior. Despite its advantages in structured decision-making, discrete action space methods remain underexplored in imitation learning (IL), where continuous control is generally favored for its ability to produce more natural driving behaviors.

\subsection{Multi-Task Learning for Imitation Learning}

Another line of research in autonomous driving incorporates multi-task learning to enhance imitation learning models by integrating classification tasks. 
In these methods, additional objectives such as high-level maneuver recognition are introduced alongside action prediction to improve generalization. 
For instance, Xu et al. \cite{xu_end--end_2017} proposed an end-to-end FCN-LSTM network that predicts both discrete motion actions and continuous control commands, using semantic segmentation as an auxiliary task to enhance environmental context understanding. Similarly, Couto and Antonelo \cite{couto2021generative} utilized a Generative Adversarial Imitation Learning framework where a discriminator network functions as a classifier, distinguishing between expert and agent-generated trajectories to guide the policy network toward expert-like behavior. Additionally, Kishky et al. \cite{kishky2024end} introduced the Conditional Imitation Co-Learning framework, enhancing interdependence between specialist network branches dedicated to specific navigational commands, and reformulating the steering regression problem as a classification task to improve performance in unseen environments. While these approaches enhance the model's ability to interpret complex driving contexts, they do not explicitly address confidence estimation at the action level, leaving uncertain predictions unflagged and uncorrected in real time.

\subsection{Uncertainty in Imitation Learning}

Uncertainty estimation in imitation learning is crucial for enhancing decision reliability, particularly in safety-critical applications like autonomous driving. While some methods focus on assessing confidence in action predictions, others address trajectory-level uncertainty in planning and prediction. This section categorizes existing work into two main areas: action uncertainty and trajectory-based uncertainty.

\textit{Action Uncertainty in End-to-End Imitation Learning:}
Embedding confidence within policy actions is often neglected in deep learning-based vehicle control models. However, advanced uncertainty estimation techniques have been explored to quantify uncertainty over control commands. Bayesian Neural Networks (BNNs) \cite{cui2019uncertainty, nozarian2020uncertainty} offer a probabilistic framework for uncertainty estimation. The Uncertainty-Aware Imitation Learning (UAIL) algorithm \cite{cui2019uncertainty} applies Monte Carlo Dropout to estimate uncertainty in control outputs, selectively acquiring new training data in uncertain states to prevent error accumulation. 
Additionally, Wang and Wang \cite{wang2024uncertainty} integrate Bayesian uncertainty estimation via mean-field variational inference (MFVI) within the Data Aggregation (DAgger) framework to mitigate covariate shift and improve model performance in unseen environments. 
Tai et al. \cite{tai2019visual} introduce an uncertainty-aware deployment strategy for visual-based autonomous driving, where a stochastic domain adaptation mechanism generates multiple image transformations to estimate aleatoric uncertainty. While effective in handling visual domain shifts, this approach introduces additional computational overhead and does not directly assess action confidence in the control policy.

\textit{Trajectory-Based Uncertainty for Planning and Prediction:}
Beyond action confidence, another line of research examines uncertainty in trajectory prediction and planning. Zhou et al. \cite{zhou2022long} focus on long-horizon planning, ensuring robustness in rare driving scenarios by leveraging a prediction uncertainty-aware trajectory planning framework that accounts for the inherent uncertainty in long-tail driving events, improving decision-making in low-data or high-risk situations. VTGNet \cite{cai2020vtgnet} introduces a stochastic trajectory generation network for urban environments, capturing multimodal motion patterns. Similarly,  Cai et al. \cite{cai2020probabilistic} employ multimodal sensor fusion to reduce trajectory uncertainty and improve navigation performance.

While these uncertainty estimation techniques improve model robustness, they often introduce significant computational complexity and require specialized training procedures. In contrast, our approach enhances the native capability of the imitation learning model to assess and express action confidence without relying on supplementary models or data aggregation. By integrating a classification head alongside the regression head within the same neural network architecture, we enable real-time confidence estimation directly from the model’s outputs. This design maintains computational efficiency and simplifies the training process, offering a more streamlined solution for robust imitation learning.

    \section{Method}
\label{sec:method}

One key limitation of imitation learning is its inability to assess confidence in its regression-based steering predictions, leading to potential safety risks in uncertain scenarios. To address this, we propose a dual-output architecture that combines continuous regression for steering angle prediction with discrete classification to estimate confidence. By evaluating classification confidence and entropy, our approach detects low-confidence situations, allowing corrective measures such as adjusting the predicted action or activating a predefined fallback mechanism such as handing control over to a human operator. This Confidence-Guided Fusion framework enhances decision reliability by dynamically identifying and mitigating uncertain predictions, ensuring more robust and stable autonomous control.

\subsection{Model Architecture}
As shown in Fig.~\ref{fig:method-intro}, the proposed model consists of an image encoder 
with two parallel output heads; a continuous regression head that predicts a raw steering angle and a discrete classification head that assigns a probability distribution over $N$ discrete steering bins in the range $[-1,1]$.

The model architecture is detailed in Table \ref{tbl:model-arch}. A ResNet-50 \cite{he2016deep} backbone pretrained on ImageNet \cite{russakovsky2015imagenet} extracts image features, with the first 120 layers frozen to retain general representations and the rest fine-tuned for the driving task. 
Given an input image $I$, the model maps it to both a continuous and a discrete steering prediction:

\begin{equation} 
    f: I \rightarrow (y_{\text{cont}}, y_{\text{disc}}), 
\end{equation}

where $y_{cont}$ is the continuous steering output and $y_{disc}$ represents the classification probabilities over discrete steering bins. 
The ResNet-50 output, a 2048-dimensional feature vector, is processed by two fully connected layers ($256 \rightarrow 128$) with ReLU activation. The regression and classification heads each contain a single fully connected layer (64 neurons, ReLU). Each fully connected layer is followed by a dropout layer with a dropout rate of $0.3$. The regression head produces $y_{cont}$ via a linear output layer, while the classification head outputs $y_{disc}$ using a fully connected layer with $N$ neurons and Softmax activation.

\begin{table}[!t]
    \scriptsize
    \renewcommand{\arraystretch}{1.2}
	\centering
	\caption{Model Architecture Details}
	\label{tbl:model-arch}
    \begin{tabular}{c|c|c|c}
    \hline \hline
        \textbf{Component} & \textbf{Layer} & \textbf{Trainable params} & \textbf{Non-trainable} \\
        \hline \hline
        {} & ResNet-50 & 17.23M & 6.37M \\
        Encoder & FC1 (256) & 524,544 & 0 \\
        {} & FC2 (128) & 32,896 & 0  \\
        \hline \hline
        Regression Head &  FC3 (64) & 65 & 0 \\
        {} & FC reg (1) & 1 & 0 \\
        \hline \hline
        Classification Head & FC4 (64) & 64N + N & 0 \\
        {} & FC class (N) & N & 0 \\
        \hline \hline
    \end{tabular}
\vspace*{-5pt}
\end{table}





\subsection{Training Strategy}
The model is trained using a multi-task learning framework with a loss function combining both regression and classification objectives. The total loss function is defined in \eqref{eq:loss}, where $\mathcal{L}_{\text{reg}}$ is the \textit{Mean Squared Error (MSE)} loss for continuous steering output, and $\mathcal{L}_{\text{class}}$ is the \textit{Sparse Categorical Cross-Entropy (SCE)} loss for steering bin classification. The weighting terms $\lambda_1$ and $\lambda_2$ control the balance between the two loss terms:

\begin{equation}
    \mathcal{L} = \lambda_1 \mathcal{L}_{\text{reg}} + \lambda_2 \mathcal{L}_{\text{class}}.
    \label{eq:loss}
\end{equation}

To ensure stable convergence, we use the Adam optimizer with an \textit{Exponential Learning Rate Decay}:

\begin{equation}
    \eta_t = \eta_0 \times \gamma^{t / T},
    \label{eq:lr-scheduler}
\end{equation}

where $\eta_0$ is the initial learning rate, $\gamma$ is the decay rate, and $T$ represents the decay steps.

\subsection{Confidence-Based Action Correction}

The proposed method evaluates the reliability of each steering prediction by analyzing three key factors. \textit{Classification confidence (\(\max p\))} represents the highest probability assigned to any discrete class in the classification output, indicating how certain the model is about its prediction.
\textit{Classification entropy (\(H\))} quantifies uncertainty in classification predictions and is computed in:  

\begin{equation} 
    H = - \sum p_i \log p_i ,
    \label{eq:class-entropy}
\end{equation}

where \( p_i \) represents the probability assigned to each discrete steering bin. A higher entropy value suggests greater uncertainty in classification. 
Lastly, \textit{Regression-Classification Alignment} evaluates whether the regression-predicted steering angle falls within the classification bin or its adjacent neighbors.


\begin{table*}[!t]
    \footnotesize

    \renewcommand{\arraystretch}{1.3}
	\centering
	\caption{Correction Strategy Based on Regression and Classification Alignment}
	\label{tbl:correction-strategy}
    \begin{tabular}{c|c|c}
        \hline \hline
        \textbf{Case} & \textbf{Condition} & \textbf{Correction Method} \\
        \hline \hline
        \textbf{Case 1} &  $\max p \geq \tau$ AND $\text{bin}(\hat{y}_r) = \text{bin}(\hat{y}_c)$ OR $\text{bin}(\hat{y}_r) = \text{bin}(\hat{y}_c) \pm 1$  & Retain regression output without modification. \\
        \hline \hline
        \textbf{Case 2} & $\max p \geq \tau$ AND $\text{bin}(\hat{y}_r) \notin \{ \text{bin}(\hat{y}_c), \text{bin}(\hat{y}_c) \pm 1 \}$ & Adjust regression output using probabilistic sampling \\
        {} & {} & from a uniform distribution:  
        $\hat{y} \sim U(\text{bin}_{\text{lower}}(\hat{y}_c), \text{bin}_{\text{upper}}(\hat{y}_c))$ \\
        \hline \hline
       \textbf{Case 3} & 
        $\max p < 0.5$ AND $H > 1.5$  
        & Retain regression output without modification. \\
        \hline \hline
        \textbf{Case 4} & $\max p < \tau$ AND $H \leq 1.5$ AND $\text{bin}(\hat{y}_r) \neq \text{bin}(\hat{y}_c)$ & Adjust regression output using probabilistic \\
        {} & {} & sampling from a normal distribution: $\hat{y} \sim \mathcal{N}(\hat{y}_r, \sigma^2)$ \\
        \hline \hline
    \end{tabular}
\end{table*}
 
These three factors combined determine the categorization of each prediction into one of four cases, as shown in Table \ref{tbl:correction-strategy}, where an appropriate correction method is applied accordingly.
Case 1 is when classification is confident and aligns with the regression bin (or neighbors), the regression output is kept. Case 2 is when classification is confident but regression falls outside the expected range, classification is prioritized, and a value is sampled within its bin to ensure smooth transitions. Case 3 is when classification is uncertain (high entropy), regression output is retained for stability. Finally, 
Case 4 is when both are uncertain but classification entropy is low, regression is refined using probabilistic sampling to enhance robustness while keeping smooth adjustments.
Based on the strategy in Table~\ref{tbl:correction-strategy}, we formalize the confidence-guided decision process into a structured algorithm. Algorithm~\ref{alg:correction} dynamically adjusts steering predictions based on classification confidence, entropy, and regression-classification alignment, ensuring consistent and reliable decision-making.

\begin{algorithm}[!h]
\small
\caption{Confidence-Based Correction Algorithm}
\label{alg:correction}
\textbf{Input:} $y_{\text{cont}}$ (Continuous output), $y_{\text{disc}}$ (Discrete output), $B$ (Bin edges), $\tau$ (Confidence threshold, default: 0.9) \\
\textbf{Output:} $y_{\text{final}}$ (Refined final output)
\begin{algorithmic}[1]
    \State $c_{\max} \gets \max(y_{\text{disc}})$
    \State $i_{\max} \gets \arg\max(y_{\text{disc}})$
    \State $i_{\text{cont}} \gets \text{digitize}(y_{\text{cont}}, B)$
    \State Compute entropy: $H \gets -\sum p_i \log (p_i + \epsilon)$
    \State Initialize $y_{\text{final}} \gets \text{NaN}$

    \If{$c_{\max} \geq \tau$ \textbf{and} $i_{\text{cont}} \in \{i_{\max}, i_{\max} + 1, i_{\max} - 1\}$}
        \State $y_{\text{final}} \gets y_{\text{cont}}$
    \ElsIf{$c_{\max} \geq \tau$ \textbf{and} $i_{\text{cont}} \notin \{i_{\max}, i_{\max} + 1, i_{\max} - 1\}$}
        \State $a \gets B[i_{\max}], b \gets B[i_{\max} + 1]$
        \State Sample $N$ values from $\mathcal{U}(a, b)$
        \State $y_{\text{final}} \gets \frac{1}{N} \sum \text{samples}$
    \ElsIf{$c_{\max} < 0.5$ \textbf{and} $H > 1.5$}
        \State $y_{\text{final}} \gets y_{\text{cont}}$
    \Else
        \State $\sigma^2 \gets \text{Var}(y_{\text{disc}})$
        \State Sample $N$ values from $\mathcal{N}(y_{\text{cont}}, \sigma^2)$
        \State $y_{\text{final}} \gets \frac{1}{N} \sum \text{samples}$ 
    \EndIf
    \State \textbf{Return} $y_{\text{final}}$
\end{algorithmic}
\end{algorithm}

\subsection{Evaluation Metrics}
To assess the performance of our action correction approach, we employ two categories of evaluation metrics: \textit{classification performance metrics} and \textit{route similarity metrics} \cite{jekel2019similarity}. Classification metrics evaluate the accuracy of steering predictions, while route similarity metrics quantify how well the corrected trajectories align with reference paths.
\subsubsection{Classification Performance Metrics}
The classification accuracy of steering decisions is evaluated using precision, recall, F1-score, and accuracy, which are widely used in supervised learning \cite{bishop2006pattern}. These metrics provide insight into the model’s ability to balance false positives, false negatives, and overall correctness in classification tasks.
\subsubsection{Route Similarity Metrics}
To measure how well the corrected trajectories align with reference paths, we use geometric and temporal similarity metrics, provided by \cite{jekel2019similarity}:

\begin{itemize}
    \item \textit{Fréchet Distance} quantifies the worst-case deviation between predicted and reference trajectories.
    \item \textit{Dynamic Time Warping (\(DTW\)) Distance} measures alignment between trajectories while allowing temporal distortions.
    \item \textit{Area Between Curves (\(ABC\))} captures cumulative spatial deviation between predicted and reference paths.
    \item \textit{Curve Length (\(CL\))} evaluates trajectory smoothness, with shorter paths indicating more stable corrections.
\end{itemize}

\subsubsection{Statistical Analysis of Trials}
Each experiment is repeated 10 times per route type to account for variability in predictions. For each metric, we report the mean value, representing overall performance, and the standard deviation (std), which indicates consistency—where lower values suggest more stable corrections.

These metrics provide a comprehensive evaluation of both the model’s predictive accuracy and its effectiveness in refining trajectories for smoother, and more robust driving.

    \section{Experimental Setup}
\label{sec:experimental-setup}

We conduct experiments using the CARLA simulator \cite{dosovitskiy2017carla}, a high-fidelity simulation environment for autonomous driving research. All experiments are performed in CARLA version 0.9.15.
\subsubsection{Data Collection}  
We collected two datasets, one in Town04 and another in Town06, each containing 300,000 RGB images with corresponding steering angles under ClearNoon weather conditions.
\subsubsection{Data Preprocessing}  
\begin{figure}
      \subfigure[Town04]{\includegraphics[width = 2.6cm, clip, trim=0 30 0 30 ]{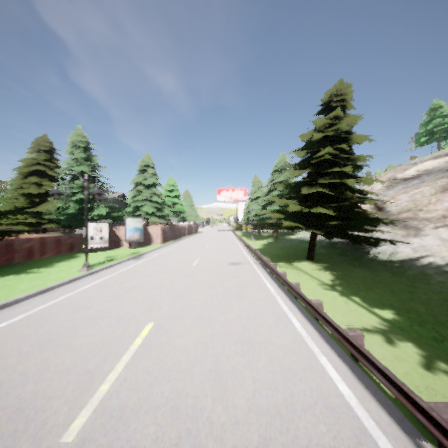}} 
      \subfigure[Town06]{\includegraphics[width = 2.6cm, clip, trim=0 30 0 30]{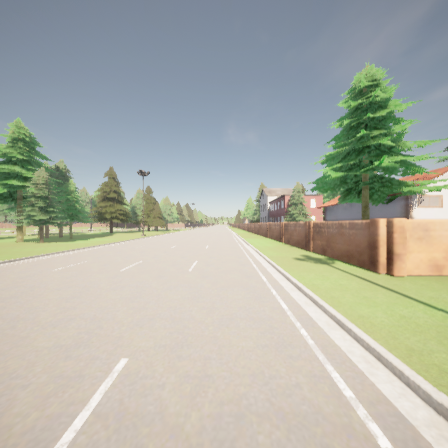}}
      \subfigure[Cropped/Resized]{\includegraphics[width = 2.6cm, clip, trim=0 20 0 20]{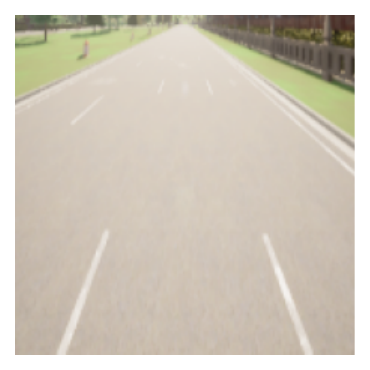}}
    \caption{Sample images of the RGB front camera.}
    \label{fig:in-img-samples}
\vspace*{-5pt}
\end{figure}

Each dataset was preprocessed separately to improve balance and model performance, resulting in $38,380$ samples from Town04 and $43,904$ from Town06. The preprocessing steps included:  
\begin{itemize}
    \item Input Images: Fig.~\ref{fig:in-img-samples} shows sample images from each town in parts (a) and (b). For computational efficiency, all images were cropped and resized to \(160 \times 160\) pixels. Part (c) of the figure illustrates the final preprocessed images used for training and testing.
    \item Normalization: Steering values were discretized into bins, and the number of samples per bin was capped to prevent the overrepresentation of certain steering angles.
    \item Augmentation: Horizontal flipping was applied, with the corresponding steering angle inverted.
    \item Steering Scaling: Angles within $[-0.25, 0.25]$ were scaled to $[-1, 1]$ by multiplying by 4.
    \item Steering Angle Discretization: Continuous steering values were mapped to $N=11$ discrete bins spanning $[-1, 1]$, ensuring structured classification. The bin size is 0.182, which makes the middle bin $[-0.091, 0.091]$.
    Each classification output represents a bin, with the final steering prediction determined by a confidence-guided fusion strategy.
\end{itemize}

\subsubsection{Training Setup}  
The dataset was divided into training and evaluation sets as follows:  
\begin{itemize}
    \item Town06 was used entirely for training to ensure a diverse dataset from a single location, minimizing distribution mismatch across train-test splits.
    \item Town04 was split ($60\%$ training, $40\%$ testing) to evaluate generalization within the same environment.
\end{itemize}
This resulted in a training dataset of 66,932 samples and a test dataset of 15,352 samples.
The model was trained using a weighted loss function \eqref{eq:loss}, where both $\lambda_1$ and $\lambda_2$ values were set to $0.5$. The learning rate followed an exponential decay schedule \eqref{eq:lr-scheduler} with initial learning rate $\eta_0 = 10^{-5}$, decay factor $\gamma = 0.96$, and decay step $T = 10,000$.

    \section{Experiments}

\begin{figure}[!t]
    \centering
    \includegraphics[width=0.3\textwidth]{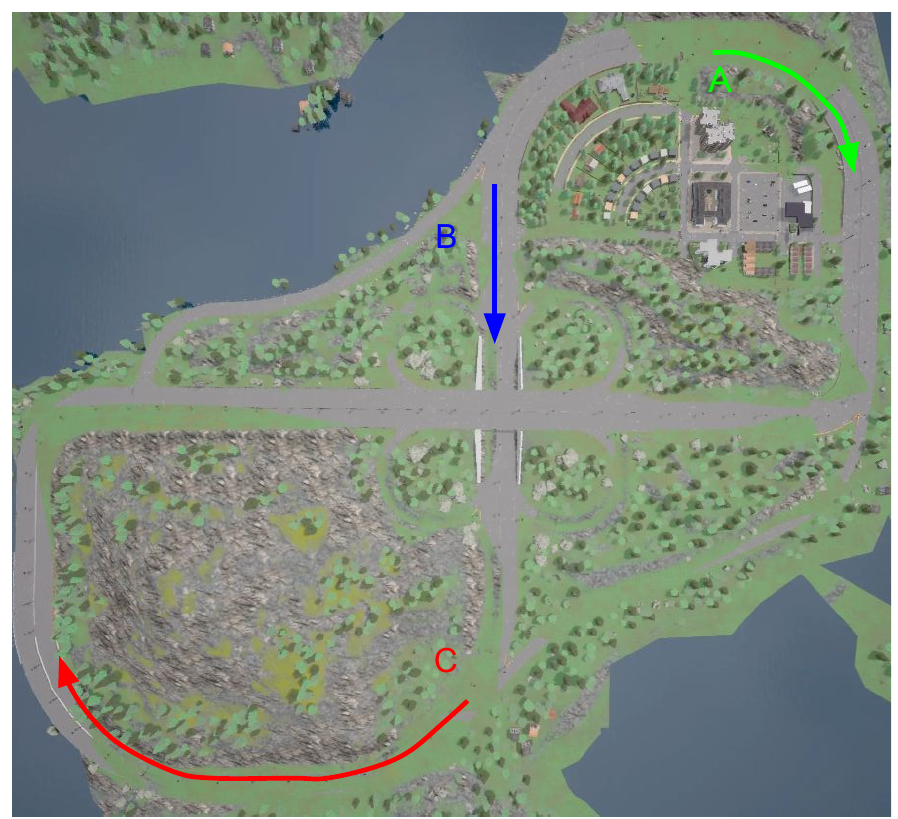}  
    \caption{Routes in CARLA's Town04 used for trajectory evaluation. Route A (green) represents a one-turn path, Route B (blue) is a straight segment, and Route C (red) is a two-turn route.}
    \label{fig:test_track}
\end{figure}

The trained model was deployed in CARLA and evaluated over 10 independent trials per route type in Town04. The test environment featured a structured road network with three predefined route types to assess the agent's driving performance. 
Each route type was defined by a fixed start and end location in Town04 to ensure consistency across all trials. The evaluation routes are shown in Fig.~\ref{fig:test_track}. To maintain uniform conditions, ClearNoon weather was used in all experiments. 
A Tesla Model 3 was used as the test vehicle, processing real-time camera inputs to predict steering angles and apply confidence-guided corrections. The control strategy included:
\begin{itemize}
    \item Steering: Continuous predictions from the model.
    \item Throttle: Binary control (0 or 0.5), adjusted based on speed.
    \item Braking: Disabled throughout the tests.
\end{itemize}

The experiments were designed to achieve two objectives:
\begin{enumerate}
    \item Demonstrate confidence estimation: By utilizing the classification head, our model provides real-time confidence distributions to assess the reliability of regression predictions.
    \item Evaluate action correction: Show how the model’s confidence-aware correction mechanism improves trajectory smoothness and enhances control robustness.
\end{enumerate}

    \section{Results}
\label{sec:results}

This section presents the evaluation of our confidence-guided model in terms of classification accuracy and route similarity analysis. First, we assess classification metrics to determine how well the model predicts steering classes. Then, we analyze the impact of confidence-guided corrections on trajectory accuracy using Fréchet distance, DTW, ABC, and CL.

\subsection{Classification Performance}

The classification results are summarized in Table~\ref{tbl:classification-results}. The model achieved an overall accuracy of ($76.33\%$), with the highest precision observed in class 5 (straight driving), aligning with dataset distribution since straight driving is the most frequent scenario. While the model performs well on common driving behaviors, it struggles with less frequent maneuvers. Notably, precision and recall for classes 4 (sharp left) and 6 (sharp right) are lower, indicating difficulty in distinguishing between sharp turns. This misclassification pattern can be attributed to class imbalance, as these classes have significantly fewer training samples. The confusion matrix in Fig.~\ref{fig:confusion_matrix} further illustrates this trend.


\begin{table}[!t]
    \scriptsize
    \renewcommand{\arraystretch}{1.2}
	\centering
	\caption{Classification Performance}
	\label{tbl:classification-results}
    \begin{tabular}{c|c|c|c|c}
        \hline \hline
        \textbf{Class} & \textbf{Precision} & \textbf{Recall} & \textbf{F1-score} & \textbf{Support} \\
        \hline \hline
        0  & 0.6469  & 0.7051  & 0.6748  & 434 \\
        1  & 0.4994  & 0.7262  & 0.5918  & 599 \\
        2  & 0.8137  & 0.6899  & 0.7467  & 1538 \\
        3  & 0.7683  & 0.6381  & 0.6971  & 956 \\
        4  & 0.4202  & 0.3241  & 0.3660  & 617 \\
        5  & 0.8738  & 0.9240  & 0.8982  & 5793 \\
        6  & 0.2998  & 0.3295  & 0.3139  & 434 \\
        7  & 0.7699  & 0.5109  & 0.6142  & 871 \\
        8  & 0.7687  & 0.8443  & 0.8047  & 1535 \\
        9  & 0.6612  & 0.6357  & 0.6482  & 571 \\
        10 & 0.6900  & 0.7117  & 0.7007  & 444 \\
        \hline \hline
        \textbf{Macro Avg} & 0.6556 & 0.6400 & 0.6415 & 13792 \\
        \textbf{Weighted Avg} & 0.7650 & 0.7633 & 0.7598 & 13792 \\
        \hline \hline
        \textbf{Accuracy} & \multicolumn{3}{c|}{0.7633} & 13792 \\
        \hline \hline
    \end{tabular}
\end{table}

\subsection{Confusion Matrix}
\begin{figure}[!t]
    \centering
    \includegraphics[width=0.8\linewidth, clip, trim=20 20 0 5]{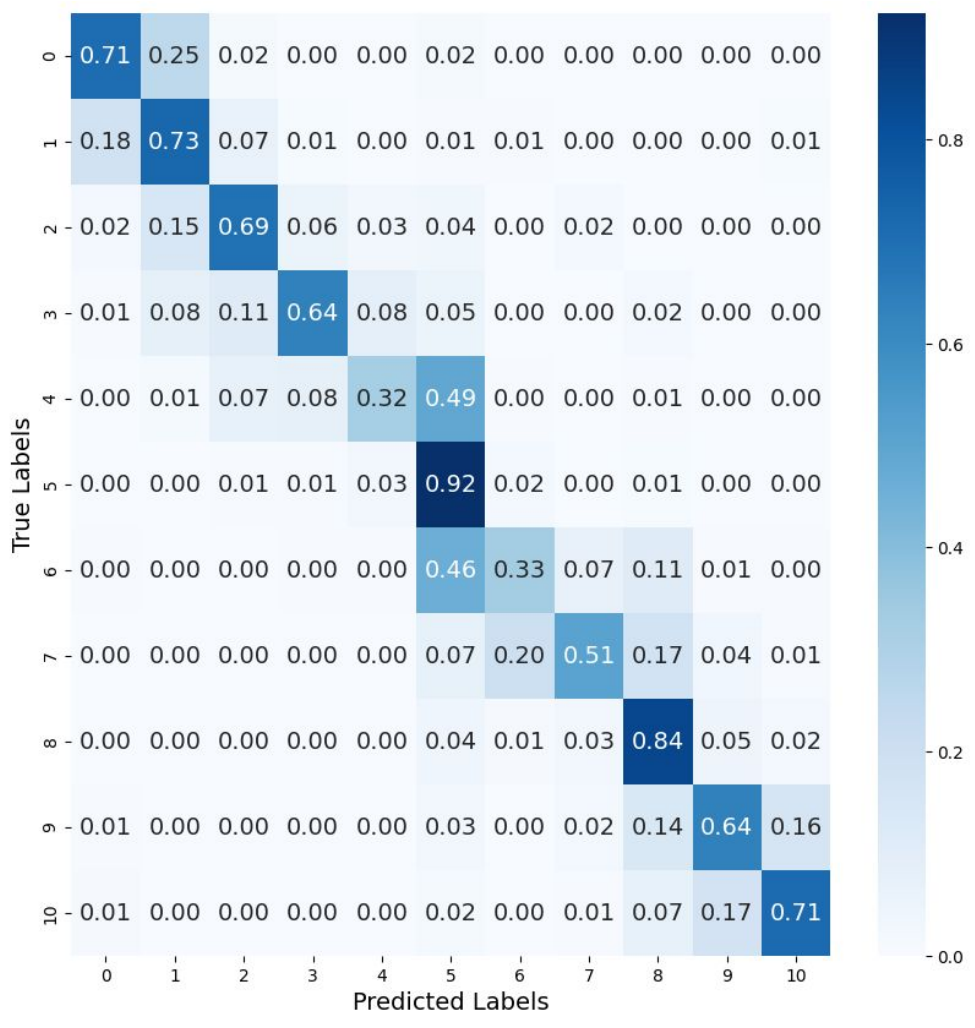}
    \caption{Confusion Matrix for the Classification Model. The \(x\) axis is for the predicted classes and the \(y\) axis is for true classes.}
    \label{fig:confusion_matrix}
\end{figure}

The confusion matrix in Fig.~\ref{fig:confusion_matrix} highlights key misclassification patterns. 
The overall misclassification rate remains within acceptable limits for real-world deployment. The largest source of error comes from sharp turns being confused with mild turns, which could lead to suboptimal steering adjustments. These results suggest that data augmentation or a weighted loss function could improve classification performance for underrepresented classes.

\subsection{Real-time Confidence Estimation}
\begin{figure}[!t]
    \centering
    \includegraphics[width=1\columnwidth, clip, trim=0 40 0 40]{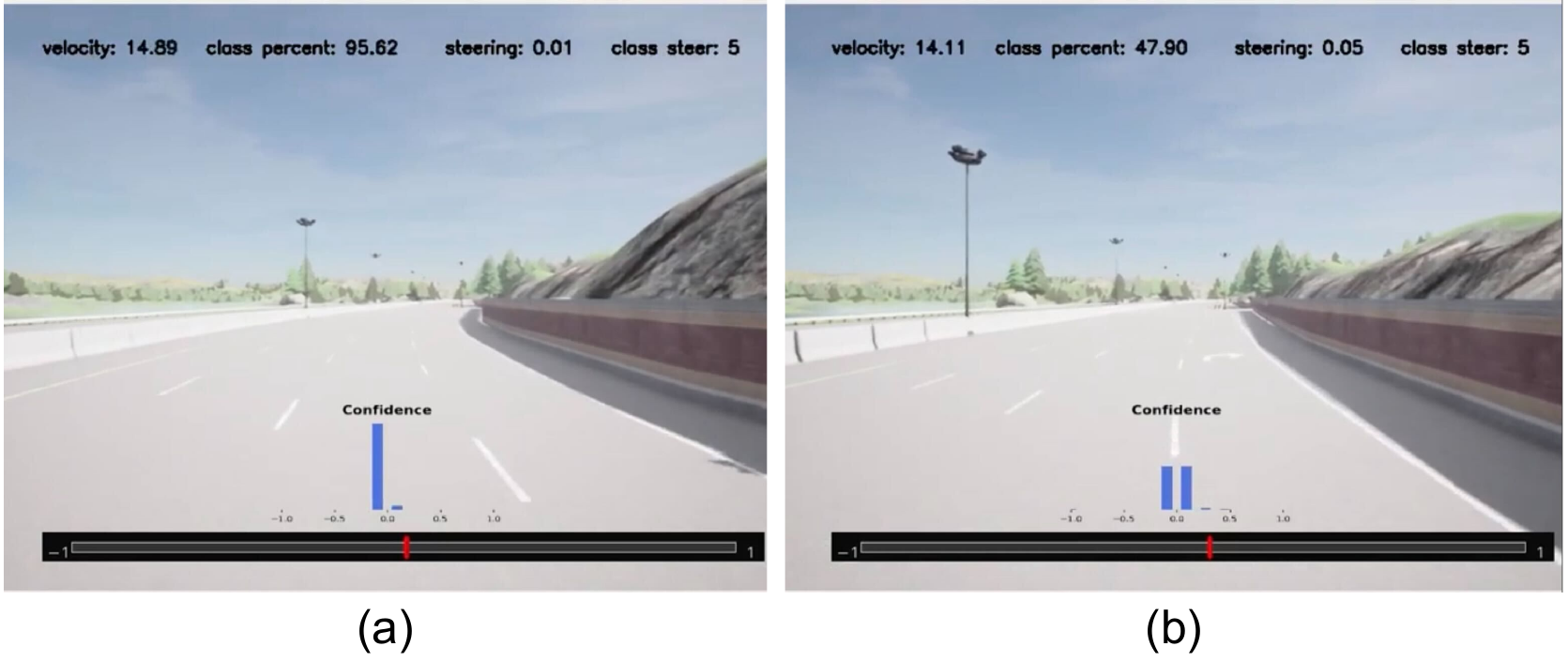}
    \caption{
             Real-time confidence estimation of the regression steering value. (Left) The vehicle remains within the lane, resulting in high confidence. (Right) As the vehicle deviates from the lane, the confidence decreases, indicating higher uncertainty in the regression output. video available: \url{https://www.youtube.com/watch?v=2RPf-T_lsLc}
            }
    \label{fig:real-time-confidence}
\end{figure}

To further assess the effectiveness of our confidence-guided model, we analyze its real-time confidence estimation capabilities. Fig.~\ref{fig:real-time-confidence} illustrates how the model evaluates the reliability of its steering predictions under different driving conditions.
In Fig.~\ref{fig:real-time-confidence}(a), the vehicle maintains a stable position within the lane, and the classification head assigns a high confidence score to the regression-based steering prediction ($95\%$). This indicates that in normal driving conditions, the model can reliably estimate steering without requiring corrections.
Conversely, in Fig.~\ref{fig:real-time-confidence}(b), the vehicle starts deviating from the lane. As a result, the classification head outputs a low confidence score ($47\%$), signaling higher uncertainty in the regression output. This drop in confidence activates the correction mechanism, allowing the model to refine its predictions and steer the vehicle back on track.
These results confirm that the classification-guided confidence mechanism accurately detects uncertainty, enabling adaptive corrections that improve driving stability.
By incorporating confidence-aware corrections, the model enhances safety and robustness, particularly in challenging driving conditions such as sharp turns or occluded environments.

\subsection{Route Similarity Analysis}
To evaluate the impact of confidence-guided corrections on driving trajectories, we compare trajectory similarity using Fréchet distance, DTW, ABC, and CL. The results, shown in Tables~\ref{tab:frechet_dtw_comparison}, \ref{tab:area_curve_length_comparison}, and \ref{tab:overall_similarity}, demonstrate that corrections significantly improve route accuracy. These results indicate that the confidence-guided corrections reduce trajectory deviation by up to ($50\%$), particularly in two-turn routes, where uncorrected paths exhibit significant drift. This suggests that integrating confidence measures into action selection can enhance the stability and accuracy of autonomous vehicle steering.

\begin{table}[!t]
\scriptsize
\centering
\caption{Fréchet and DTW Metrics With/Without Correction}
\label{tab:frechet_dtw_comparison}
\begin{tabular}{lcc|cc}
\toprule \toprule
\textbf{Route Type} & \multicolumn{2}{c|}{\textbf{Fréchet Mean ↓}} & \multicolumn{2}{c}{\textbf{DTW Mean ↓}} \\
 & \textbf{W/ Corr.} & \textbf{W/O Corr.} & \textbf{W/ Corr.} & \textbf{W/O Corr.} \\
\midrule
One-Turn  & \textbf{6.6643}  & 7.1122  & \textbf{1108.42}  & 1148.90  \\
Straight  & \textbf{1.4979}  & 1.5609  & \textbf{145.27}   & 160.82   \\
Two-Turn  & \textbf{8.9296}  & 25.9943 & \textbf{5417.38}  & 6605.88  \\
\midrule \midrule
 & \multicolumn{2}{c|}{\textbf{Fréchet Std ↓}} & \multicolumn{2}{c}{\textbf{DTW Std ↓}} \\
\midrule
One-Turn  & 1.1817  & \textbf{0.6102}  & 214.69   & \textbf{62.25}    \\
Straight  & \textbf{0.1532}  & 0.1606  & 14.28    & \textbf{13.23}    \\
Two-Turn  & \textbf{1.5184}  & 36.2325 & \textbf{1292.96}  & 3242.37  \\
\bottomrule \bottomrule
\end{tabular}
\end{table}
\begin{table}[!t]
\scriptsize

\centering
\caption{ABC and Curve Length Metrics With/Without Correction}
\label{tab:area_curve_length_comparison}
\begin{tabular}{lcc|cc}
\toprule \toprule
\textbf{Route Type} & \multicolumn{2}{c|}{\textbf{Area Mean ↓}} & \multicolumn{2}{c}{\textbf{Curve Length Mean ↓}} \\
 & \textbf{W/ Corr.} & \textbf{W/O Corr.} & \textbf{W/ Corr.} & \textbf{W/O Corr.} \\
\midrule 
One-Turn  & \textbf{635.16}  & 655.99  & \textbf{0.2259}  & 0.2361  \\
Straight  & \textbf{32.26}   & 41.56   & \textbf{0.3506}  & 0.5032  \\
Two-Turn  & \textbf{3267.10} & 3726.18 & \textbf{0.5956}  & 1.4787  \\
\midrule \midrule
 & \multicolumn{2}{c|}{\textbf{Area Std ↓}} & \multicolumn{2}{c}{\textbf{Curve Length Std ↓}} \\
\midrule 
One-Turn  & 105.47  & \textbf{37.82}   & \textbf{0.0455}  & 0.0164  \\
Straight  & \textbf{9.71}    & 8.45    & 0.1291  & \textbf{0.1458}  \\
Two-Turn  & \textbf{401.67}  & 1097.17 & \textbf{0.0585}  & 1.9527  \\
\bottomrule \bottomrule
\end{tabular}
\end{table}
\begin{table}[!t]
\scriptsize
\centering
\caption{Overall Route Similarity With/Without Correction}
\label{tab:overall_similarity}
\begin{tabular}{lcc|cc}
\toprule \toprule
\textbf{Metric} & \multicolumn{2}{c|}{\textbf{Mean ↓}} & \multicolumn{2}{c}{\textbf{Std ↓}} \\
 & \textbf{W/ Corr.} & \textbf{W/O Corr.} & \textbf{W/ Corr.} & \textbf{W/O Corr.} \\
\midrule 
Fréchet               & \textbf{5.6973}  & 11.5558  & \textbf{3.3410}  & 22.8185  \\
DTW                   & \textbf{2223.69} & 2638.54  & \textbf{2443.11} & 3401.99  \\
ABC  & \textbf{1311.51} & 1474.58  & \textbf{1447.18} & 1749.69  \\
CL          & \textbf{0.3907}  & 0.7393   & \textbf{0.1769}  & 1.2186   \\
\bottomrule \bottomrule
\end{tabular}
\end{table}

    \section{Conclusion and Future Work}
\label{sec:conclusion}

In this work, we introduced a dual-head action prediction framework for imitation learning-based autonomous driving, integrating both regression and classification to enhance decision reliability. Our approach leverages the classification head to estimate confidence in regression-based steering predictions, enabling real-time uncertainty assessment and action correction. Experimental results demonstrated that confidence-guided corrections improve trajectory accuracy by up to ($50\%$).
While our model enhances interpretability and robustness through confidence estimation, we do not claim to achieve the best driving performance. More sophisticated architectures or higher-quality datasets could further improve control precision. However, our goal is to highlight the benefits of integrating classification with regression rather than purely optimizing driving performance. Future work could explore alternative uncertainty modeling techniques, adaptive loss weighting, or integration with reinforcement learning to further refine the action correction mechanism.

By addressing the fundamental limitation of confidence estimation in imitation learning, this work highlights the potential of dual-head architectures in making imitation learning autonomous driving systems more reliable, interpretable, and adaptable to uncertain environments.

    \bibliographystyle{unsrt}
\bibliography{3_end/bibliography.bib}
\end{document}